\newcommand{\CLASSINPUTbottomtextmargin}{54pt}
\newcommand{\CLASSINPUTtoptextmargin}{54pt}
\newcommand{\CLASSINPUTinnersidemargin}{54pt}
\newcommand{\CLASSINPUToutersidemargin}{54pt}

\documentclass[letterpaper, 10pt, conference]{IEEEtran}
\IEEEoverridecommandlockouts
\renewcommand{\IEEEtitletopspace}{20pt}
\usepackage[subpreambles=true]{standalone}
\usepackage{svg}
\usepackage{cite}
\usepackage{amsmath,amssymb,amsfonts}
\usepackage{dsfont}
\usepackage[ruled,linesnumbered,commentsnumbered]{algorithm2e}
\usepackage{graphicx}
\usepackage{textcomp}
\usepackage{xcolor}
\usepackage{xcolor-solarized}
\usepackage{setspace}
\usepackage{import}
\usepackage{mathtools}
\usepackage{enumitem}
\usepackage{float}
\usepackage[switch]{lineno}
\usepackage{multirow}
\usepackage{makecell}
\usepackage{xcolor}
\usepackage{caption}
\usepackage{booktabs}
\usepackage{hyperref}
\usepackage{amsthm}
\usepackage{tikz}

\def\BRULEARXIV{1}

\def\BibTeX{{\rm B\kern-.05em{\sc i\kern-.025em b}\kern-.08em
    T\kern-.1667em\lower.7ex\hbox{E}\kern-.125emX}}

\newcommand{\LEPD}{LEPD}
\newcommand{\LEPDs}{LEPDs}
\newcommand{\R}{\mathbb{R}}
\newcommand{\Pbb}{p}
\newcommand{\1}{\mathds{1}}

\DeclareMathOperator*{\argmin}{argmin}
\newcommand{\figurespace}{\vspace{-10pt}}
\renewcommand{\baselinestretch}{0.98}

\begin{document}
\title{Belief Roadmaps with Uncertain\\Landmark Evanescence
\thanks{This research was supported by the Office of Naval Research under Contract W911NF-17-2-0181, the Low Cost Autonomous Navigation \& Semantic Mapping in the Littorals program and their support is gratefully acknowledged. Any opinions, findings, conclusions or recommendations expressed in this material are those of the authors and do not necessarily reflect the views of our sponsors.}
}

\author{Erick Fuentes$^{1}$, Jared Strader$^{2}$, Ethan Fahnestock$^{1}$, 
 Nicholas Roy$^{1}$
\thanks{$^{1}$ E. Fuentes, E. Fahnestock, and N. Roy are with the Computer Science and Artificial Intelligence Laboratory, Massachusetts Institute of Technology, Cambridge, MA 02139 USA (email: {erick, ekf, nickroy}@csail.mit.edu)}
\thanks{$^{2}$ J. Strader is with the Laboratory for Information and Decision Systems, Massachusetts Institute of Technology, Cambridge, MA 02139 USA (e-mail: jstrader@mit.edu)}
 }
\maketitle

\if\BRULEARXIV 1
\begin{tikzpicture}[overlay, remember picture]
\path (current page.north)
node[align=center, below=8pt] {This article has been accepted for publication in the IEEE Conference on Robotics and Automation \\
Please cite as: Erick Fuentes, Jared Strader, Ethan Fahnestock, Nicholas Roy, \\ "Belief Roadmaps with Uncertain Landmark Evanescence", in IEEE Conference on Robotics and Automation};    
\end{tikzpicture}
\fi

\begin{abstract}
We would like a robot to navigate to a goal location while minimizing state uncertainty.
To aid the robot in this endeavor, maps provide a prior belief over the location of objects and regions of interest.
To localize itself within the map, a robot identifies mapped landmarks using its sensors.
However, as the time between map creation and robot deployment increases, portions of the map can become stale, and landmarks, once believed to be permanent, may disappear.
We refer to the propensity of a landmark to disappear as landmark evanescence.
Reasoning about landmark evanescence during path planning, and the associated impact on localization accuracy, requires analyzing the presence or absence of each landmark, leading to an exponential number of possible outcomes of a given motion plan.
To address this complexity, we develop BRULE, an extension of the Belief Roadmap.
During planning, we replace the belief over future robot poses with a Gaussian mixture which is able to capture the effects of landmark evanescence.
Furthermore, we show that belief updates can be made efficient, and that maintaining a random subset of mixture components is sufficient to find high quality solutions.
We demonstrate performance in simulated and real-world experiments.
Software is available at \href{https://bit.ly/BRULE}{\textcolor{blue}{\underline{https://bit.ly/BRULE}}}.

\end{abstract}

\section{Introduction}

To reliably reach a goal location, a robot must plan in real-time with imperfect knowledge of the robot and world states.
A map, created from a previous deployment or another information source (e.g. overhead imagery), is often used as an aid in navigation.
However, the fidelity of the map tends to degrade as the time between map creation and robot deployment increases.
For example, in an urban environment, a parked car may be a useful landmark over a time span of minutes, but almost certainly ceases to be useful over days.
To the robot, the landmark simply disappeared. We refer to the propensity of landmarks to disappear as \emph{landmark evanescence}. 
The simultaneous localization and mapping (SLAM) community has studied landmark evanescence~\cite{rosen_towards_2016, nobre_online_2018, bateman_better_2020} and developed models of the behavior of these landmarks.
While the use of maps is common, the incorporation of landmark evanescence probability distributions (\LEPDs{}) into navigation models has not yet been explored.

\begin{figure}[t] 
    \centering
    \includegraphics[width=\linewidth]{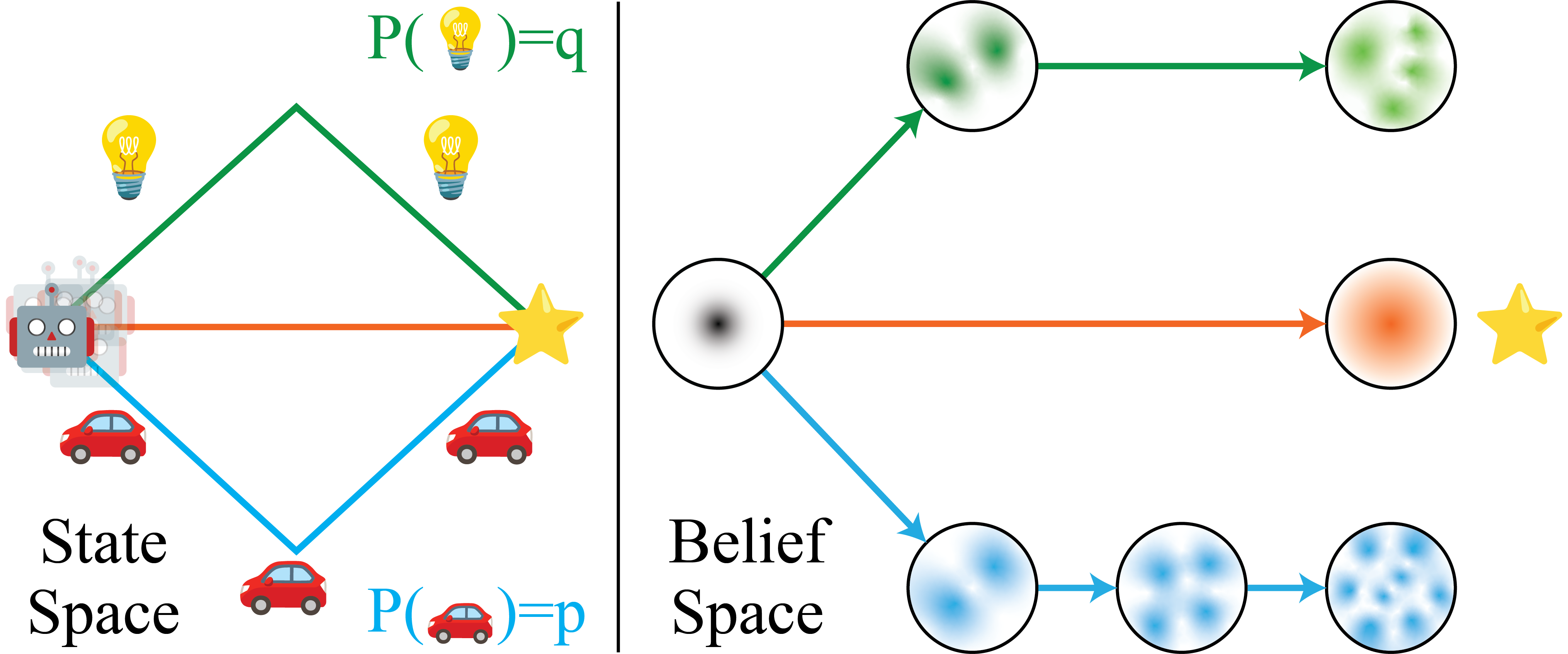}
    \caption{A robot, uncertain of its own position, must navigate to the goal.
    There is a direct path, in orange, from the start to the goal location.
    However this path does not contain any landmarks, so the robot's position uncertainty becomes large.
    The green and blue paths have evanescent landmarks that may or may not be present, leading to many possible position estimates and uncertainties along these paths.
    An upper bound on the number of uncertainties is exponential in the number of landmarks in the environment, which is untenable, even in simple environments.
    In this work, we show how to tractably plan with landmark evanescence.
    } 
    \label{fig:hero}
    \figurespace{}
\end{figure}

Consider the situation in Fig.~\ref{fig:hero}, where a robot, with some initial uncertainty, must navigate to a goal.
The robot has access to a map which contains the locations of lamp posts and parked cars.
Lamp posts are detectable when illuminated, which occurs with probability $q$.
Parked cars are detectable if they have not moved, which occurs with probability $p$. 
During planning, the robot can simulate how its belief would evolve after moving or observing a landmark.
The robot could take the orange path which proceeds directly to the goal, or one of the longer green or blue paths.
The orange path has no landmarks to aid in localization, so the position uncertainty becomes large.
Along the green path, two lamp posts maybe illuminated, resulting in four possible landmark presence configurations and a different set of robot position distributions for each configuration.
Along the blue path, three parked cars may be present, resulting in eight possible presence configurations and position distributions.
In general, the number of position distributions along a path is exponential in the number of landmarks in the environment.
When choosing a path, ignoring landmark evanescence risks brittle navigation, but considering landmark evanescence requires care to avoid the inherent combinatorial complexity.

A general way of modeling this problem is as a Partially Observable Markov Decision Process (POMDP)~\cite{kaelbling_planning_1998}.
However, in practice, POMDP solvers struggle to produce good solutions in the presence of continuous state or action spaces, or long horizons\cite{pineau_anytime_2006,sunberg_online_2018}.
The Belief Roadmap (BRM) ~\cite{prentice_belief_2009} can provide a solution to our problem, but only when the landmarks are known to be present. Other efforts have considered how to handle the cases of intermittent observations~\cite{bopardikar_robust_2016}, or how to handle uncertainty in obstacles~\cite{missiuro_adapting_2006}.
However, none of these works tackle the problem of landmark evanescence and the resulting hybrid discrete-continuous belief.

In this work, we introduce an approach for belief space planning with landmark evanescence. 
We show that a Gaussian mixture can approximate the belief over the robot pose and landmark evanescence. We derive an efficient belief update and extend the BRM to Gaussian mixture beliefs.
We show that despite the exponential number of configurations of landmark presence, a random subset of mixture components is sufficient to find quality solutions.
We demonstrate the approach in simulated and real-world experiments.

\section{Problem Formulation}
\label{sec:problem_formulation}

Given a model of our robot dynamics and sensors, and a prior map, we wish to navigate from a start location to a goal location while maximizing the probability that we are near our goal.
We assume access to a landmark evanescence probability distribution (\LEPD{}) that describes how the world may have evolved since it was mapped\footnote{The \LEPD{} can come from a data driven method, as described by \cite{rosen_towards_2016, nobre_online_2018, bateman_better_2020}, or it can be derived from expert knowledge. If one assumes a static world, this can also be represented as \LEPD{}}.
We describe each of these components in the following subsections.

\subsection{World Model}
\label{sec:world_model} 

Let a landmark be $\ell \in \R^d \times I$, where $d$ is the dimension of landmark locations and $I$ is a set of identifiers. 
Each landmark is uniquely labeled with an identifier so that landmarks and measurements may be associated.
A map $M = \{\ell_i\}_{i=1}^{N_l}$ is a collection of landmarks. 
The set of binary strings of length $N_l,\,\Omega = \{0, 1\}^{N_l}$ is used to represent the presence or absence of the mapped landmarks, which we refer to as an \textit{evanescence configuration}, where the $i$th landmark in $\omega \in \Omega =\{\omega_1, \dots, \omega_{N_l}\}$ is present if $\omega_i=1$. 
We define a probability measure $\Pbb(\omega): 2^{\Omega} \rightarrow [0,1]$ that maps a set of configurations to a probability. 
We assume that the evanescence configuration is fixed during the deployment.

\subsection{Robot Model}
\label{sec:robot_model}
Let $X \subset \R^{d_X}$ be the $d_X$-dimensional robot state space and let $U \subset \R^{d_U}$ be the $d_U$-dimensional control space.
If $x_t\in X$ and $u_t\in U$ are the state of the robot and control applied at time $t$, then the next state $x_{t+1}$ is distributed according to the distribution $\Pbb(x_{t+1} | x_t, u_t )$.
Let $Z=(\R^{d_Z} \cup \{o\}) \times I$ be the observation space where $d_Z$ is the dimension of a measurement, $o$ represents the absence of a measurement, and $I$ is the set of landmark identifiers.
Concretely, for every landmark, the robot either receives a measurement ($z\in \R^{d_Z}$) or notes a lack of a measurement ($z = o$), with perfect data association of measurements to mapped landmarks.
Let $\omega \in \Omega$ be an evanescence configuration. The probability of receiving a measurement $z_{i, t}\in Z$ of the $i$th landmark at time $t$ is given by $\Pbb(z_{i,t} | x_t, \omega)$.
During planning, we assume a perfect detector that produces range and bearing measurements when the distance to the landmark is less than $r_{max}$ and produces $o$ otherwise\footnote{We assume the landmark detector is noise free during planning. During execution, if we are also estimating which landmarks are present, it would be important to model noise in the landmark detections.}.

\subsection{Bayesian Update of Belief}
If we allow a Markov assumption to be made, future states depend only on the current state and control, that is $\Pbb(x_{t+1} | x_{0:t}, u_{0:t}, z_{1:t}) = \Pbb(x_{t+1} | x_t, u_t)$.
Additionally, the current observations depend only on the current state and evanescence configuration, that is $\Pbb(z_t | x_{0:t}, u_{0:t}, z_{1:t-1}, \omega) = \Pbb(z_t | x_t, \omega)$.
Let $b_{t} = \Pbb(x_{t}, \omega | u_{0:t-1}, z_{1:t})$ represent our belief of the robot state and evanescence configuration at time $t$.
Then the belief at the next time step $b_{t+1}$ can be computed recursively using the Bayes filter\cite{thrun_probabilistic_2005} update:
\begin{align}
\begin{split}
b_{t+1|t} &= \Pbb(x_{t+1}, \omega | u_{0:t}, z_{1:t}) = \\
&\qquad\int_{x_{t} \in X} \Pbb(x_{t+1} | x_{t}, u_{t}) b_{t} dx_{t} \label{eqn:bayes_process_update}
\end{split} \\
b_{t+1} &= \frac{1}{\eta}\Pbb(z_{t+1} | x_{t+1}, \omega) b_{t+1|t} \label{eqn:bayes_measurement_update}
\end{align}
where $\eta$ is a normalization constant. Equations~\eqref{eqn:bayes_process_update}~and~\eqref{eqn:bayes_measurement_update} are known as the process and measurement update respectively. Let the combined update be given by $b_{t+1}=\tau(b_t, u_t, z_{t+1})$.

\subsection{Trajectory Planning in Belief Space}

The most common formulation of trajectory planning minimizes a cost $J: X^T \times U^T \rightarrow \R$.
In the presence of uncertainty, we might wish to minimize the expected cost.
However, defining the cost explicitly as a function of the belief allows the use of both decision theoretic and information theoretic costs.
Therefore, we wish to find a sequence of controls $u_{0:T-1}$ that minimizes an objective $J': \mathbb{B}^T \times U^T \rightarrow \R$, where $\mathbb{B}$ is the set of possible beliefs. 
Restricting to Markovian costs, the objective has the form:
\begin{equation}
    J'(b_{0:T}, u_{0:T-1}) = c_T(b_T) + \sum_{t=0}^{T-1} c_t(b_t, u_t).
\end{equation}
where $c_t$ is a per timestep cost and $c_T$ is a terminal cost.
We wish to solve the following optimization: 
\begin{equation}
\begin{aligned}
\argmin_{u_{0:T-1}}\; & J'(b_{0:T}, u_{0:T-1}) \\
\textrm{s.t.} \quad & b_0 = b[0] \\
& b_{t+1} = \tau (b_t, u_t, z_{t+1})
\label{eqn:optimization_problem}
\end{aligned}
\end{equation}

In this work, we examine the case where we wish to minimize uncertainty at the goal, and allow for potentially high uncertainty beliefs along the path, so long as the uncertainty is resolved when the robot reaches the goal. However, the approach is not inherently limited by this choice.

\section{Belief Roadmaps}

POMDP solvers could be used to solve \eqref{eqn:optimization_problem}.
However, the optimization quickly becomes intractable as the planning horizon increases\cite{pineau_anytime_2006}, so we instead base our planning approach on the belief roadmap (BRM)\cite{prentice_belief_2009}.
The BRM navigates to a goal while minimizing uncertainty given a dynamics model, an observation model, and a map of the environment.
In this section, we revisit the major components of the BRM and identify the challenges introduced by landmark evanescence.

\textbf{Gaussian Belief:} To make the problem tractable, the BRM makes a number of assumptions.
First, the BRM restricts the possible trajectories to a graph where the nodes represent locations and the edges encode a trajectory between nodes, shortening the apparent planning horizon with macro-actions, and enabling computational reuse.
Next, the BRM assumes that the initial uncertainty is well modeled by a Gaussian and that the dynamic and observation models are linearizable.
These assumptions simplify the updates in \eqref{eqn:bayes_process_update} and \eqref{eqn:bayes_measurement_update} into those of an Extended Kalman Filter (EKF)\cite{thrun_probabilistic_2005}.
The BRM determinizes the observations by assuming the maximum likelihood observation is received.
As a result, there is no error in the EKF measurement update, making the evolution of the state distribution deterministic.
For our problem, the belief is no longer Gaussian.
To remedy this shortcoming, we use a Gaussian mixture belief where each component captures the belief under a given evanescence configuration.

\textbf{Efficient Belief Update:} The BRM runs a breadth first search, propagating beliefs across an edge using the EKF updates for each node expansion.
Since the EKF is a recursive filter, traversing an edge with different initial beliefs normally requires applying each update in turn.
However, \cite{prentice_belief_2009} finds that an appropriate factorization of the belief covariance allows the updates to be gathered into a single step, greatly accelerating the search.
For our problem, a different belief means the fast updates are not applicable. We show that with some bookkeeping, the fast updates can be recovered.

\textbf{Search Pruning:} If each node in the graph has a branching factor $b$, then the number of paths of length $\ell$ is $b^\ell$. 
The BRM uses a dominance check to prune branches of the search tree, reducing the number of considered paths.
Specifically, if a path reaches a node that was traversed by another path with lower uncertainty, then the first path is pruned.
In \cite{prentice_belief_2009}, the trace of the covariance matrix is used to compare two beliefs.
For our problem, we must define how to compare two Gaussian mixtures to enable pruning. We use the  probability mass within a region to perform this comparison.

\section{Belief Roadmaps with Uncertain Landmark Evanescence}

In this section, we present two attempts at reconciling landmark evanescence with the assumptions of the BRM.
The first, which we call Belief Roadmaps with Uncertain Landmark Evanescence (BRULE), replaces the belief used in the BRM with a Gaussian mixture.
An alternative, which we call BRULE-Expected (BRULE-E), resolves the inconsistencies by applying the BRM on configurations sampled from the \LEPD{} and then evaluating the generated paths.

\subsection{Gaussian Mixture Belief Representation}

A Rao Blackwellized belief\cite{koller_probabilistic_2009} is one where a subset of the variables are sampled and the remaining variables are described by a distribution conditioned on the sampled variables. Note that conditioned on an evanescence configuration, the robot state belief is Gaussian. 
Therefore, a belief can be used where each evanescence configuration $\omega$ has a \textit{particle belief} $p_{\omega}(x) = \Pbb(x | \omega)$ with an associated \textit{particle weight} $\Pbb(\omega)$. The resulting belief is a Gaussian mixture:
\begin{align}
\Pbb(x) = \sum_{\omega \in \Omega} \Pbb(\omega) N(x | \mu_{\omega}, \Sigma_{\omega}).
\end{align}

Since $|\Omega| = 2^{|M|}$, the number of particles required to represent the belief exactly becomes intractable, even in simple environments.
However, there is an opportunity for efficiency gains.
Initially, when no observations have been made, all particles have the same particle belief.
After the robot observes landmark $\ell_i$, the particles $P_0=\{p_\omega \mid \omega_i = 0\}$ make one update and the particles $P_1=\{p_\omega \mid \omega_i = 1\}$ make a different update.
We refer to $P_0$ and $P_1$ as consistent sets of particles.
If $n$ landmarks have been observed, we expect there to be $2^n$ consistent sets.
The particle weight is the total probability of the associated consistent set.
Since $n \leq |M|$, significant savings can be realized by instead associating each consistent set with a particle. However, the number of particles still grows exponentially, which we address in \ref{sec:bound_size_of_belief}.

\subsection{Belief Update with Landmark Evanescence}

Next we tackle the propagation of the new belief across an edge.
We start with the general process and measurement updates shown in \eqref{eqn:bayes_process_update} and \eqref{eqn:bayes_measurement_update} and show that closed form updates can be derived using the existing BRM assumptions.
 Starting with the process update, we incorporate the evanescence configuration $\omega$:
\begin{equation}
b_{t+1 | t} = \int_{x_t} \Pbb(x_{t+1}| x_t, u_t) \underbrace{\Pbb(x_t, \omega | u_{0:t-1}, z_{1:t})}_{b_t} dx_t.\end{equation}
Taking advantage of independence relations and rearranging:
\begin{align}
\begin{split}
b_{t+1 | t} &= \Pbb(\omega | u_{0:t-1}, z_{1:t}) \\ &\quad\underbrace{\int_{x_t} \Pbb(x_{t+1}| x_t, u_t) \Pbb(x_t | \omega, u_{0:t-1}, z_{1:t}) dx_t}_{\textrm{EKF Process Update}}. \label{eqn:mixture_process_update}
\end{split}
\end{align}
We see that each particle is updated by maintaining the particle weight and then performing the usual EKF process update to the associated particle belief.
 
Now we define the measurement update.
We take advantage of the law of total probability to split the observations of the landmark $\ell_i$ into two cases, one where a measurement is acquired ($z \in \R^{d_Z}$) and one where it is not acquired ($z=o$),
\begin{align}
\begin{split}
\Pbb(z | x_{t+1}, \omega) &= \Pbb(\1_{z = o} | x_{t+1}, \omega)\Pbb(z | \1_{z = o}, x_{t+1}, \omega) \\
&\quad + \Pbb(\1_{z \neq o} | x_{t+1}, \omega)\Pbb(z | \1_{z \neq o}, x_{t+1}, \omega),
\end{split}
\end{align}
where $\1_{z=o}$ is the indicator function that equals one when $z=o$. The measurement update for $\ell_i$ then becomes:
\begin{align}
b_{t+1} &= \frac{1}{\eta} b_{t+1 | t} \Pbb(z_{i,t+1} \mid x_{t+1}, \omega) \\
\begin{split}
&= \frac{1}{\eta}[\underbrace{\Pbb(\1_{z_{i,t+1} = o} | x_{t+1}, \omega) \Pbb(\omega | u_{0:t}, z_{1:t})}_{A} \\
&\underbrace{\Pbb(z_{i,t+1} | \1_{z_{i,t+1} = o}, x_{t+1}, \omega) \Pbb(x_{t+1} | \omega, u_{0:t}, z_{1:t})}_{B} \\
&+ \underbrace{\Pbb(\1_{z_{i,t+1} \neq o} | x_{t+1}, \omega)\Pbb(\omega | u_{0:t}, z_{1:t})}_{C} \\
&\underbrace{\Pbb(z_{i,t+1} | \1_{z_{i,t+1} \neq o}, x_{t+1}, \omega)\Pbb(x_{t+1} | \omega, u_{0:t}, z_{1:t})}_{D}] \label{eqn:mixture_measurement_update}
\end{split}
\end{align}
where terms $A$ and $C$ correspond to the updated particle weights depending on whether $\ell_i$ is observed or not, and terms $B$ and $D$ are the updated particle beliefs.

Let us consider a few cases to gain a better understanding of the update.
When $\omega_i=0$, then $\forall x_{t+1}\in X,\Pbb(\1_{z_{i,t+1} \neq o}=1 | x_{t+1} \omega) = 0$, so the term $CD=0$, and only the term $AB$ remains.
Since $\Pbb(\1_{z_{i,t+1}=o}=1|x_{t+1}, \omega) = 1$, the particle weight remains unchanged.
Additionally, $\Pbb(z_{i,t+1}=o|\1_{z_{i,t+1}=o}, x_{t+1}, \omega) = 1$ and the particle belief also remains unchanged.
Therefore, if $\ell_i$ is known to be absent, the associated observation updates do not change the belief.

Measurement updates when $\omega_i=1$ require more care.
Under the observation model in Sec. \ref{sec:robot_model}, a lack of measurement ($z_{i,t+1}=o$) implies that the true state must be at least $r_{max}$ away from $\ell_i$.
Since the particle belief has infinite support, some portion of the belief will be within detection range and the rest will lie outside.
Incorporating this non-Gaussian observation would break our Gaussian particle belief assumption.
This non-Gaussian property greatly complicates the estimation problem, but in planning, we can choose to leverage the maximum likelihood observation assumption to condition the observations on the mean of the particle belief $\mu_{t+1}$ instead of on $x_{t+1}$.
Therefore, when $\ell_i$ is more than $r_{max}$ away from $\mu_{t+1}$, we assume that no measurement is received and the particle belief is unchanged, as above. When $\ell_i$ is less that $r_{max}$ away from $\mu_{t+1}$, the entire belief is updated with the received observation. In this case, we recognize $D$ as an EKF measurement update. We also note that all particle beliefs share the mean.

Using \eqref{eqn:mixture_process_update} and \eqref{eqn:mixture_measurement_update}, we can efficiently predict future beliefs to help in solving \eqref{eqn:optimization_problem}.
The Gaussian mixture belief combined with the maximum likelihood observation assumption means that we can propagate each particle using the efficient BRM updates developed in \cite{prentice_belief_2009}.

\subsection{BRM Pruning Heuristic with Gaussian Mixture Belief}

The BRM uses a pruning heuristic to limit the number of paths under consideration. Specifically, if a path visits a node that was previously visited and the new path has higher uncertainty (as measured by the trace of the covariance) than the previous visit, then the new path is pruned.
As the belief is now a Gaussian mixture, a new method of computing a scalar quantity of uncertainty is required\footnote{There are a myriad of methods that could be devised.
For example, one could consider computing the entropy of the distribution, or computing the trace for each particle belief and weighting by the particle weight.
Bopardikar et al\cite{bopardikar_robust_2016} show that an upper bound of the maximum eigenvalue has useful properties. One could also consider the belief trajectory, such as the maximum uncertainty.}, $||\cdot||_{\Pbb}: \mathbb{B} \rightarrow \R^+$.
In this work, we use the probability mass within a region $R \subset X$ around the mean state belief, although the effect of different norms could be interesting and is left for future work. Concretely, we define:
\begin{align}
||b||_{\Pbb} = E_{\omega \sim \Pbb(\omega)}\left[\int_{x \in R} \Pbb(x\mid \omega) dx\right]
\end{align}

With these modifications, we are now able to run the BRM with a Gaussian mixture belief.
However, as was previously noted, the number of particles in a belief grows exponentially in the number of landmarks observed along a path.
While this computational complexity may be manageable in environments with sparse landmarks, it becomes untenable in even modestly dense environments. 
We now show how to bound the number of particles.

\subsection{Bounded Size of Belief}
\label{sec:bound_size_of_belief}

To determine the relative value of one path over another, BRULE only requires access to the belief through the size of the uncertainty $||b||_{\Pbb}$.
We propose tracking an approximate belief $\tilde{b}$, which maintains a bounded number of particles, chosen in such a way that $||\tilde{b}||_{\Pbb}\approx||b||_{\Pbb}$.

Let $b$ be a belief with $N$ particles.
The approximate belief $\tilde{b}_n$ with $n$ particles is constructed by sampling $I_n\subset \{1, \dots, N\}$ where $|I_n|=n$ without replacement according to the particle weights.
In \cite{murphy_rao-blackwellised_2001}, it is proved that estimates of expectations based on a particle filter converge almost surely with error governed by the central limit theorem.
While this result was derived assuming sampling with replacement, in \cite{hoeffding_probability_1963} and \cite{serfling_probability_1974}, it was shown that estimates when sampling without replacement have lower error. 

The naive implementation of weighted sampling without replacement is inefficient.
The method proposed in~\cite{efraimidis_weighted_2006} samples in $\mathcal{O}(n \log k)$ time where $n$ is the number of particles and $k$ is the number of desired samples.
Since the particle beliefs are not required when sampling particles, significant time and space is saved by deferring the computation of edge transforms and updated beliefs until after sampling.

\subsection{BRULE-Expected}

An alternative approach to solving the optimization problem in \eqref{eqn:optimization_problem}, is to sample paths and select the path that performs best in expectation.
The performance of this method relies heavily on the ability to focus sampling on high quality paths.
We propose sampling evanescence configurations $\{\omega^i\}_{i=1}^{N}, \omega^i \sim \Pbb$ and then using the conventional BRM to find a path $p^i$ for each sampled configuration $\omega^i$, and then using the path that performs best in expectation over all sampled configurations.
The paths sampled are of high quality because each path is optimal for some configuration. However, even in the limit, the optimal path over all possible configurations is not guaranteed to be sampled\footnote{Consider an environment with two landmarks. Let the \LEPD{} be such that at most one of the landmarks is present. The optimal path may visit both potential landmarks locations, but since BRULE-E generates candidate plans from sampled configurations, a path that visits both landmarks cannot be generated. This scenario is further examined in the real-world experiments.}.


\section{Evaluation}

To evaluate the proposed BRULE and BRULE-E algorithms, we perform a series of simulation and real-world experiments.
The simulation experiments are aimed at understanding the impact of the number of particles on the uncertainty reduction and computational cost.
The real-world experiments are aimed at validating the problem formulation.

\subsection{Simulation Experiments}

\begin{figure}
\centering
\includesvg[width=\linewidth]{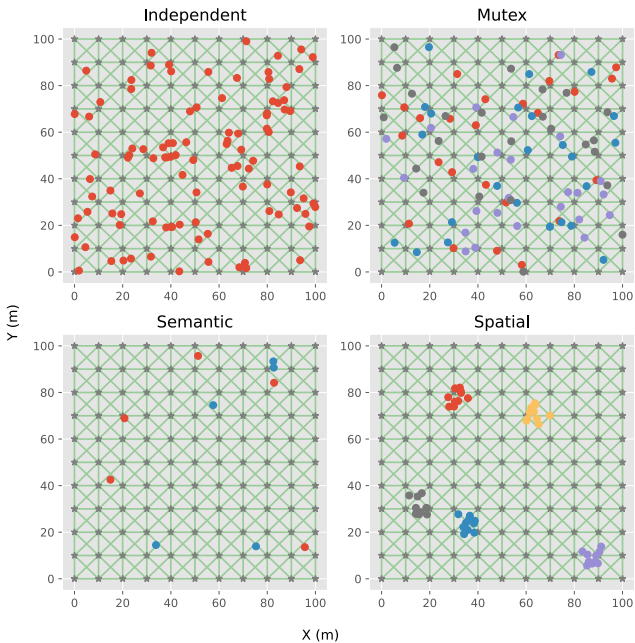}
\caption{Four example environments are shown above. The roadmap nodes and edges are shown in dark gray and green respectively. The colored dots are landmarks. All landmarks of the same color are correlated and are independent of differently colored landmarks, except for the environment in the top left where all landmarks are independent.}
\label{fig:sim_environments}
\figurespace{}
\end{figure}

To evaluate the proposed approach, we perform simulated experiments in a 100 m $\times$ 100 m region.
We use a regular 8-connected grid with 10 m spacing as the roadmap.
For each experiment, an environment consisting of landmark locations and an \LEPD{} are sampled.
The roadmap, landmarks, and \LEPD{} are used by each planner to create a plan.
The expected probability mass near the goal is computed by rolling out each plan using 1000 evanescence configuration samples.

As a baseline, we consider the original BRM with an optimistic assumption that all previously mapped landmarks are present.
For the BRULE and BRULE-E algorithms, we evaluate the performance as the number of particles maintained or paths sampled increases.
As the difficulty of environments can vary dramatically, we compute the regret against a planner that has privileged information.
Specifically, for each of the evaluation trials, we run the BRM with the sampled configuration.
Note that the planner with privileged information may still perform worse since pruning performed during search may discard a lower uncertainty plan.

We explore two kinds of correlation structures, the \emph{mutex} structure and the \emph{latent} structure.
In the mutex structure, exactly one landmark in each set of landmarks is present.
In the latent structure, there is a latent variable $z \sim \textrm{Bern}(p_z)$.
If $z=0$, then no landmarks are present.
If $z=1$, then each landmark is present with probability $p_l$, and is independent of every other landmark.
Completely independent landmarks can be achieved by setting $p_z=0$.
We also explore two kinds of spatial landmark distributions, diffuse and clustered.
Diffuse landmarks are sampled uniformly from the simulation region without any constraints.
Clustered landmarks are sampled in a two step process.
First, cluster centers are sampled uniformly.
Next, landmarks are sampled uniformly from a 10 m $\times$ 10 m region around the cluster center.
We sample environments by combining these correlation structures and spatial distributions.
We explore the behavior of the different methods in 66 different environments which are described in Table \ref{tab:Simulation Environment Types}.
We show examples of the environments in Fig. \ref{fig:sim_environments}.

\begin{figure}
\centering
\includesvg[width=\linewidth]{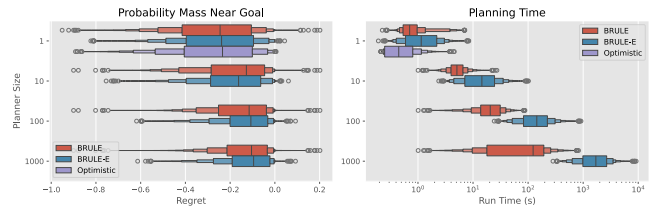}
\caption{In the left plot, a box plot of the regret is shown across the maximum number of particles maintained by BRULE and the number of samples for BRULE-E. Performance tends to improve as the number of samples used increases. In the right plot, we see the run time associated with each of the methods. We see that as the size of the planners increases, BRULE has comparable performance to BRULE-E with a greatly reduced runtime.}
\label{fig:sim_results_across_size}
\end{figure}

\begin{table}
    \centering
    \caption{Simulation Environment Types}
    \label{tab:Simulation Environment Types}

    \begin{tabular}{cccc}
    \toprule
    Name & Spatial Dist. & Evanescence Dist. & Count \\
    \midrule
    Independent & Diffuse & Latent ($p_z=0$) & 10 \\
     Mutex & Diffuse & Mutex  & 6 \\
     Semantic & Diffuse & Latent & 30 \\
     Spatial & Clustered & Latent & 20 \\
    \bottomrule
    \end{tabular}
    \figurespace{}
\end{table}

The results for the simulation experiments are shown in Fig.~\ref{fig:sim_results_across_size} with 13200 trials per planner.
We see that even with 10 particles or samples, BRULE and BRULE-E outperform the optimistic BRM baseline. 
As the size of the planners increase, we see that performance improves and they remain comparable with each other. However, BRULE scales more favorably in runtime as the number of samples increases, nearly an order of magnitude at 1000 samples. 

\subsection{Real-World Experiments}

\begin{figure}
    \centering
    \begin{center}
    \begin{tikzpicture}[outer sep=0,inner sep=0]
    \node (trajectories) {\includesvg[width=\linewidth]{figures/trajectories.svg}};
    \node (spot) at (-2.97,-0.97)
    { \includegraphics[clip,trim={7cm 10cm 15cm 5cm},height=0.9cm]{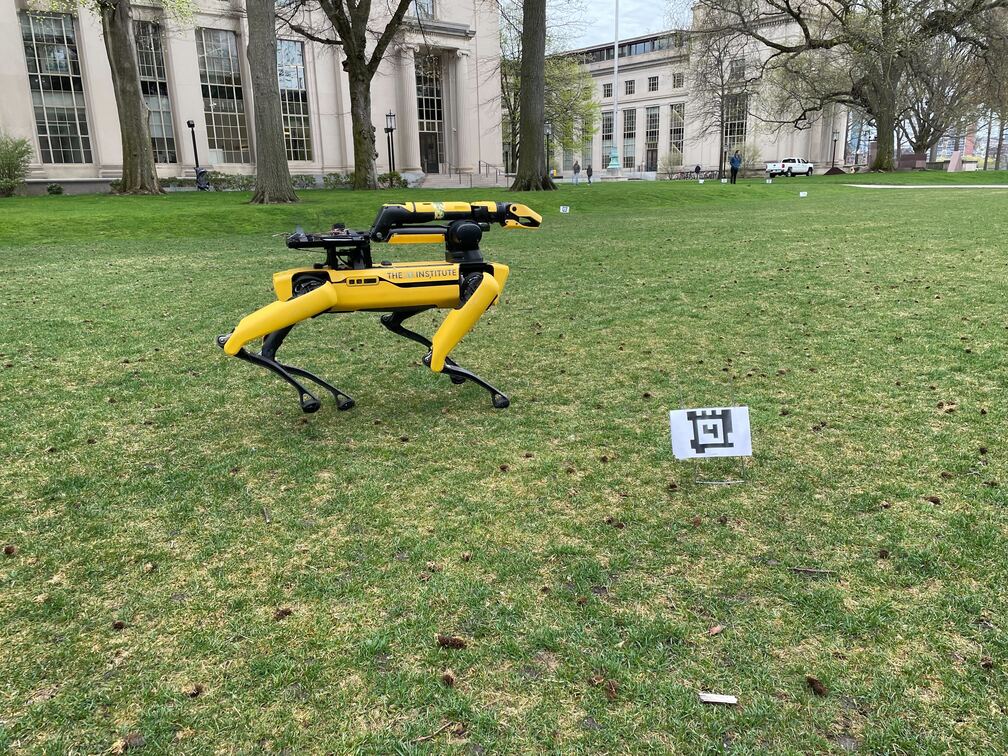}};
    \end{tikzpicture}
    \end{center}
    \vspace{-10pt}
    \caption{On the left, the test environment for real world experiments using a Boston Dynamics Spot. A mutex distribution is prescribed for landmarks 6 and 9. All other landmarks are marked absent. On the right, six executions of each plan computed by the two methods are shown. Three trials have the landmark 6 present and three trials have the landmark 9 present. Due to the mutex correlation, BRULE-E fails to recover the path that visits both landmarks. However, BRULE reasons about the correlation and determines that a longer path minimizes uncertainty at the goal.}
    \label{fig:spot_experiment_trajectories} 
\end{figure}

\begin{figure}
    \centering
    \includesvg[width=\linewidth]{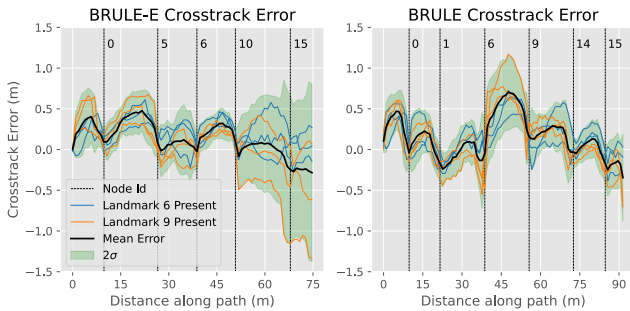}
    \caption{The crosstrack error from the nominal path for each planner.
    The more informative path discovered by BRULE yields a tighter spread of executions.}
    \label{fig:spot_experiment_cross_track_error}
    \figurespace{}

\end{figure}

To validate our problem formulation, we perform a limited set of trials in MIT's Killian Court on a real robot.
We manually place landmarks, prescribe an \LEPD{}, define a roadmap, and ask the robot to navigate to the goal location while minimizing uncertainty (see left plot in Figure \ref{fig:spot_experiment_trajectories}).
We use the Boston Dynamics Spot as the robot platform, which provides $360^\circ$ camera coverage, odometry measurements, and a low level navigation stack that accepts body relative pose commands.
All image processing, state estimation, and planning is performed on an Nvidia Xavier NX.
To create the landmarks, we place AprilTags\cite{wang_apriltag_2016} in the environment and use the \texttt{apriltag-ros}\footnote{\href{https://wiki.ros.org/apriltag_ros}{\textcolor{blue}{\underline{https://wiki.ros.org/apriltag\_ros}}}} node\cite{brommer_long-duration_2018, malyuta_long-duration_2020} to get range and bearing measurements in the robot frame from the camera images.
The odometry is fused with landmark range and bearing observations using an EKF SLAM filter\cite{sola_simulataneous_2014} to maintain a belief over the landmark locations and robot pose.
This belief, along with the specified \LEPD{} and roadmap, serve as the inputs for planning.
Instead of physically removing landmarks, we configure the estimator to ignore detections that come from absent landmarks.

In this experiment, we prescribe a mutex distribution over the landmarks near node 6 and node 9.
The landmarks near the start are always present so the robot starts with a well known belief.
All other landmarks are set to absent.
The BRULE and BRULE-E planners search for a plan to minimize uncertainty at the goal.
Each trajectory is executed six times, three times with landmark 6 present and three times with landmark 9 present.
To compute the ground truth trajectory, the estimator is rerun offline while consuming all detections.
Crosstrack error is computed by discretizing the nominal path at 1 m intervals, finding the nearest pose, and then computing the lateral offset of the nearest pose.

In the right plot of Fig.~\ref{fig:spot_experiment_trajectories}, since BRULE-E only has sample access to the \LEPD{}, we observe that it fails to correctly reason about the mutual exclusion of landmarks and plans a path that only visits one of the two potential landmarks.
This leads to the higher crosstrack error seen in Fig. \ref{fig:spot_experiment_cross_track_error} as half of the trajectories do not observe any landmarks.
In contrast, the Gaussian mixture belief maintained by BRULE allows the planner to produce a path that visits both landmarks, resulting in a smaller crosstrack error.
BRULE achieves a mean position error of 55 cm and a standard deviation of 8 cm, while BRULE-E achieves a mean error of 69 cm and a standard deviation of 51 cm.
We again note the tighter standard deviation of the BRULE trials as compared to the BRULE-E trials. The mean error for both is similar and includes error from odometry and mapping.

\section{Related Works}
Localization and Mapping in changing environments has been an active area of research.
The concept of a \LEPD{} was studied by Rosen et. al.~\cite{rosen_towards_2016}, who used the concept of hazard functions from survival analysis to describe the evanescence of a landmark.
Nobre et. al.~\cite{nobre_online_2018} extends \cite{rosen_towards_2016} to introduce correlations between landmarks and show how a \LEPD{} can be used to improve data association.

Since the original publication of the Belief Roadmap \cite{prentice_belief_2009}, additional sources of uncertainty have been incorporated into the framework or other search strategies have been developed.
The Robust BRM\cite{bopardikar_robust_2016} considers the problem tackled by the BRM with the additional complexity that sensor observations are intermittently available, for example ranging information from a radio beacon in the presence of occluders.
A key assumption is that absence or presence of a measurement is independent over time.
In our problem, the absence or presence of a measurement is directly correlated with whether or not the landmark is present.
Missiuro and Roy\cite{missiuro_adapting_2006} consider the case where there is no uncertainty in robot position, but the uncertainty is in obstacle position.
In \cite{indelman_towards_2016}, Indelman et al. relax several assumptions, including the need to discretize actions and the maximum likelihood observation assumptions, but they do not handle hybrid beliefs.
In \cite{van_den_berg_lqg-mp_2011}, Van den Berg et al. assumes knowledge of the LQR controller to compute beliefs using an LQG controller to find the best path among a set of candidates.
However, as we have shown in this work, finding high quality paths often requires reasoning about the correlation structure of the \LEPD{}.
FIRM \cite{agha-mohammadi_firm_2014} use the belief stabilizing property of the LQG controller to create independence between edges.
However, they assume that all states are observable at all times. 

\section{Conclusion}

In this work, we introduced the problem of landmark evanescence.
We have also shown that tracking the evolution of a richer Gaussian mixture belief can be made tractable by maintaining a subset of mixture components.
We have shown through our experiments that the approach can scale to moderately complex environments and is efficient enough for use in a real-world robotic system.
In the future, interesting avenues may include incorporating other types of environment evolution, such as landmark drift, and incorporating noisy observation models that model phenomena like occlusion.

\section{Acknowledgements}

The authors thank the MIT SuperCloud and Lincoln Laboratory Supercomputing Center for providing HPC resources and the AI Institute for providing a Spot robot for the physical experiments reported here.

\bibliographystyle{ieeetr}
\bibliography{references}

\end{document}